\documentclass[runningheads,a4paper]{llncs}

\usepackage{amssymb}
\usepackage{graphicx}

\usepackage{url}
\urldef{\mailsa}\path|{alfred.hofmann, ursula.barth, ingrid.haas, frank.holzwarth,|
\urldef{\mailsb}\path|anna.kramer, leonie.kunz, christine.reiss, nicole.sator,|
\urldef{\mailsc}\path|erika.siebert-cole, peter.strasser, lncs}@springer.com|

\usepackage{epsfig}
\usepackage{amsmath}
\usepackage{amssymb}
\usepackage{booktabs}
\usepackage{breakcites}
\usepackage{caption}
\usepackage{subcaption}
\def\onedot{. }
\def\eg{\emph{e.g}\onedot} 
\def\ie{\emph{i.e}\onedot}

\def\wrt{w.r.t\onedot} 

\newcommand{\mymin}{\operatornamewithlimits{min}}

\newcommand{\myargmax}{\operatornamewithlimits{argmax}}

\begin{document}

\mainmatter  

\title{Graph-Boosted Attentive Network for Semantic Body Parsing}

\titlerunning{Graph-Boosted Attentive Network for Semantic Body Parsing}

%
%
%
\authorrunning{Tinghuai Wang, Huiling Wang}

\author{%
	Tinghuai Wang[0000-0002-7863-3516]\inst{1}\thanks{wang.tinghuai@gmail.com}  \and 
	Huiling Wang\inst{2}
}%
\institute{
	Nokia Technologies, Finland
	\and
	Tampere University, Finland}


%
%

\toctitle{Lecture Notes in Computer Science}
\tocauthor{Authors' Instructions}
\maketitle

\begin{abstract}
	Human body parsing remains a challenging problem in natural scenes due to multi-instance and inter-part semantic confusions as well as occlusions. This paper proposes a novel approach to decomposing multiple human bodies into semantic part regions in unconstrained environments. Specifically we propose a convolutional neural network (CNN) architecture which comprises of novel semantic and contour attention mechanisms across feature hierarchy to resolve the semantic ambiguities and boundary localization issues related to semantic body parsing. We further propose to encode estimated pose as higher-level contextual information which is combined with local semantic cues in a novel graphical model in a principled manner. In this proposed model, the lower-level semantic cues can be recursively updated by propagating higher-level contextual information from estimated pose and vice versa across the graph, so as to alleviate erroneous pose information and pixel level predictions. We further propose an optimization technique to efficiently derive the solutions. Our proposed method achieves the state-of-art results on the challenging Pascal Person-Part dataset.
\end{abstract}

\section{Introduction}

Human semantic part segmentation, \ie assigning pixels with semantic class labels corresponding to the belonging human body parts, is a fundamental task in computer vision which provides richer structural representation for higher-level tasks such as video surveillance \cite{luo2013pedestrian,WangRLWK16,wang2017,WangW18,li2018pose}, person identification \cite{ma2011human,li2018diversity,kalayeh2018human}, image/video retrieval \cite{yamaguchi2015retrieving,wang11,hu2013markov,wei2017glad}, fine-grained recognition \cite{wang2010video,wang2011stylized,wang2014graph,zhang2014part}, image editing \cite{wang2015weakly,wang2015robust} and artistic rendering \cite{kyprianidis2012state,wang2013learnable,WangHC14}. 


Significant progress has been made recently on human part segmentation \cite{wang2015joint,chen2016deeplab,chen2016attention,liang2016semantic,xia2016zoom,jiang2016detangling,xia2016zoom,xia2017joint,li2017holistic} due to the advancement of deep neural networks and availability of large scale dataset with body part annotations \cite{chen2014detect}. However, there are several limitations of the existing methods which lead to failures in parsing unusual pose or interacting multi-person body parts in natural scenes. Those methods either take a bottom-up semantic segmentation approach relying on pixel-level training  \cite{chen2016deeplab,chen2016attention,liang2016semantic,jiang2016detangling,xia2016zoom}, or top-down approach incorporating person level recognition \cite{li2017holistic}, segmentation \cite{wang2015joint} or pose skeleton \cite{xia2017joint} into the unary potentials of a CRF framework. 

\begin{figure*}[t!]
	\centering
	\includegraphics[width=0.99\linewidth]{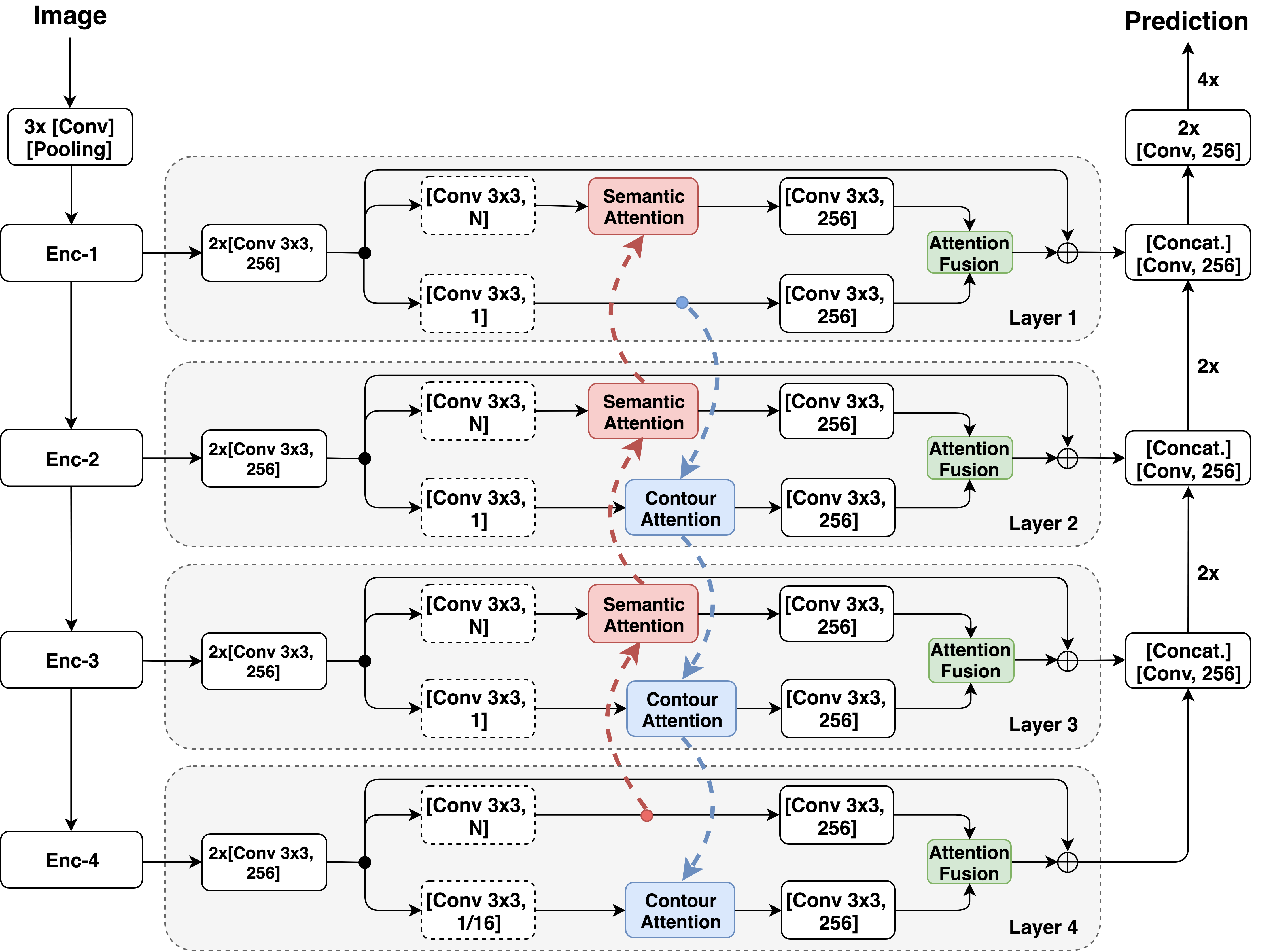}
	\caption{Illustration of the proposed network architecture.  \label{fig:net}}
\end{figure*}

The bottom-up approaches lack overall human structure and configuration information, and thus usually produce erroneous predictions when limbs intertwined, or bodies partially occluded. Some of the top-down approaches complement the bottom-up methods with object level information to constrain the pixel labeling, whereas the object level detection or segmentation suffer from multi-person scenes where people are overlapping or in unusual non-pedestrian-like body poses. As a well studied representation of people,  \cite{xia2017joint} directly uses the estimated pose skeleton as part score map which, however, suffers from early commitment --- if the pose estimation fails there is no recourse to recovery. 

Graphical models have become influential tools in computer vision \cite{wang2010multi,qi20173d,wang2017submodular,wang2019zero,wang2019graph,yang2021learning}, providing a versatile framework for representing and analyzing intricate visual data. These methods utilize the inherent structure and relationships within images \cite{wang2015robust,tinghuai2016method,xing2021learning} and videos \cite{wang2010video,wang2014wide,chen2020fine}, allowing for more advanced and context-aware analysis. By depicting visual elements as nodes and their interactions as edges, graphical models can capture spatial, temporal, and semantic dependencies essential for various applications such as visual information retrieval \cite{hu2013markov}, stylization \cite{WangCSCG10,wang2011stylized,wang2013learnable}, object detection \cite{WangW16,tinghuai2016apparatus,zhao2021graphfpn}, scene understanding \cite{WangW14,wang2017cross,wang2020spectral,tinghuai2020watermark,deng2021generative}, and image or video segmentation \cite{wang2015weakly,wang2016semi,wang2016primary,tinghuai2017method,tinghuai2018method1,ZhuWAK19,zhu2019cross,tinghuai2020semantic,lu2020video,wang2021end}.

To thoroughly exploit higher-level human configuration information in a principled manner, we propose a novel graphical model
to jointly model and infer the human body parsing problem in totally unconstrained scene scenarios, where there are large variations of people in scale, location, occlusion, and pose. 
Our approach is able to 
systematically integrate the top-down part configuration information and bottom-up semantic cues into a unified model.

The key contributions of this work are as follows.
\begin{itemize}
	\item We propose a novel architecture to overcome the fundamental obstacles to FCN based semantic body parsing, i.e., semantic ambiguity and boundary localization.
	\item We propose a novel graphical model which integrates higher-level human configuration information with local semantic cues in a principled manner.
	\item The proposed graphical model is able to recursively estimate the part configuration cues from local semantic cues and vice versa. In this manner, long-range intrinsic structure facilitates propagation of local semantic cues across higher-level body configuration. 
	\item We develop an optimization technique to simultaneously estimate the proposed two quadratic cost functions which are supplementary to each other in the proposed graph.
\end{itemize}

	\begin{figure}[!t]
	\centering
	
	\begin{subfigure}[b]{0.5\textwidth}
		\includegraphics[width=\linewidth,height=9cm]{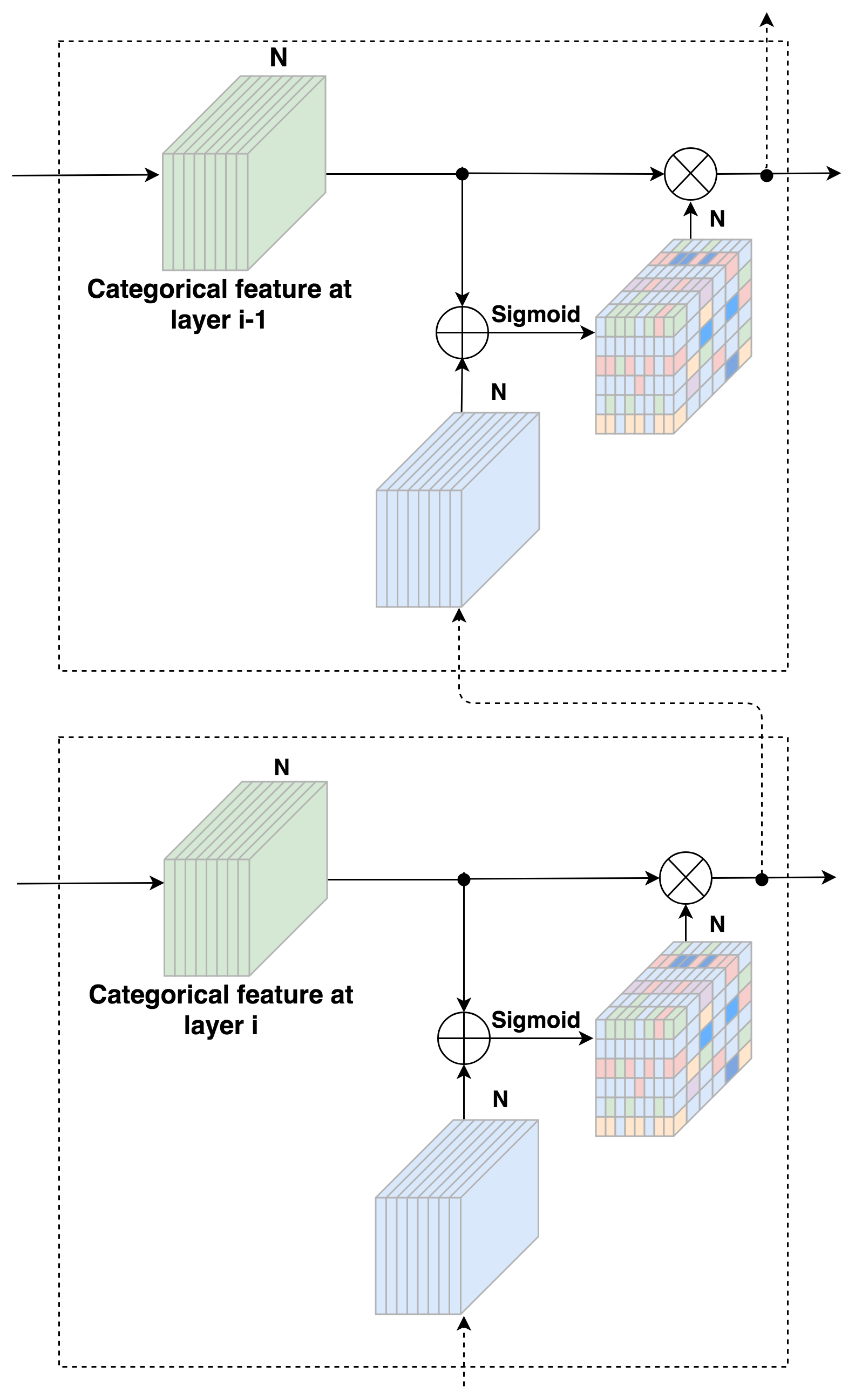}
		\label{fig:sem}
	\end{subfigure}%
	\hfill 
	\begin{subfigure}[b]{0.5\textwidth}
		\includegraphics[width=\linewidth,height=9cm]{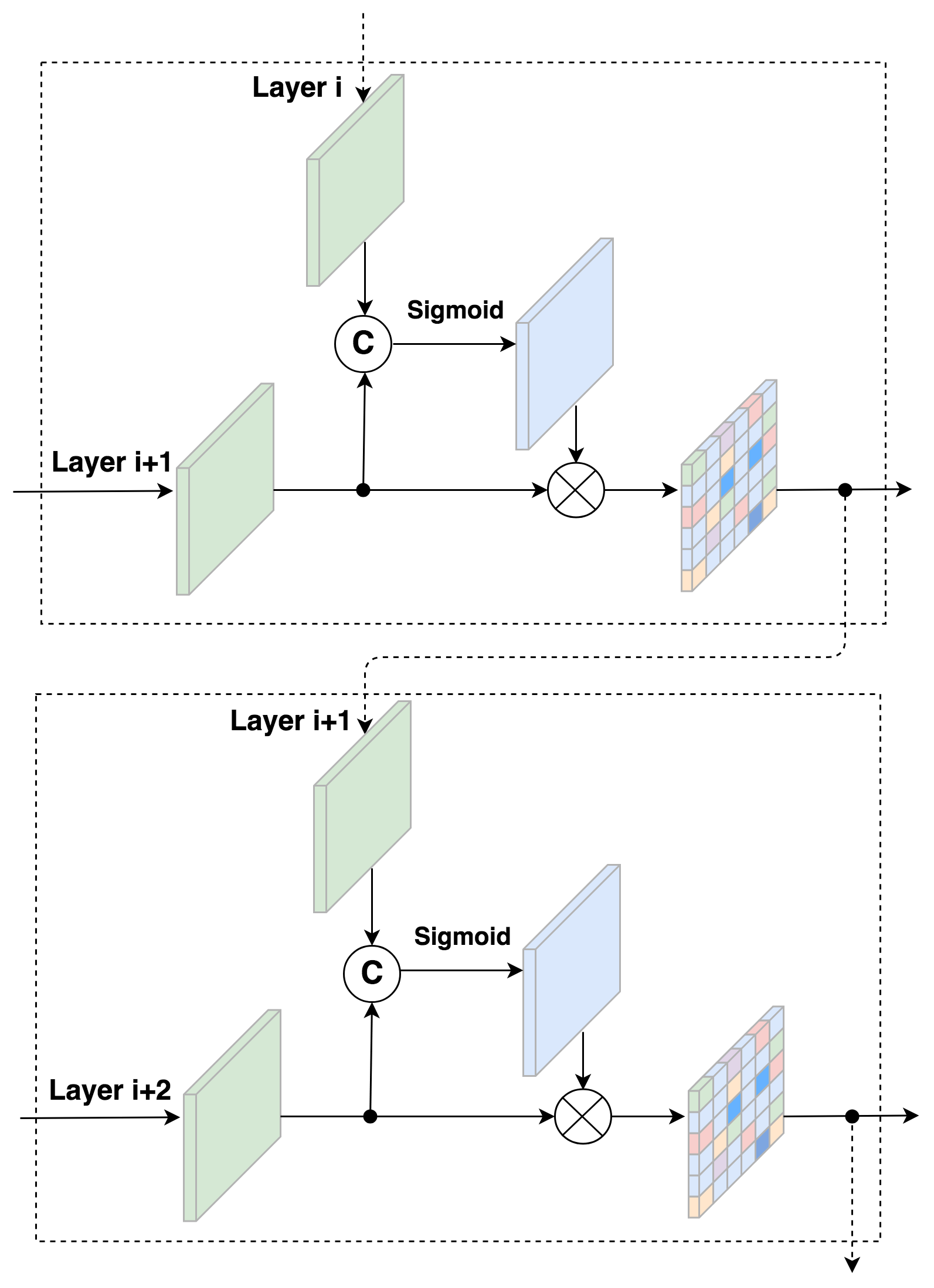}
		\label{fig:boundary}
	\end{subfigure}%
	\caption{Illustration of the proposed semantic attention module (left) and contour attention module (right). }
	\label{fig:sem-boundary}
\end{figure}

%

\section{Our Approach}

Given a $H\times W\times 3$ input image $\mathrm{I}$, our goal is to output a pixel-wise part segmentation map. We start by designing and training a network to predict a stack of initial probability maps $H\times W\times N$ where $N=P+1$ and $P$ is the number of semantic part classes with one background class. 

\subsection{Network Architecture}

The overall network architecture is illustrated in Fig. \ref{fig:net}, which is an encoder-decoder network.  The encoder consists of four blocks, i.e., \emph{Enc-i} (i=1,2,3,4), according to 
the size of the feature maps. Encoder extracts feature maps with decreasing spatial resolutions (due to pooling and strided convolution) and increasing semantic information in the forward pass. As a consequence, feature maps from early layers retain fine visual details, i.e., boundary related information, and little semantic knowledge, whilst the coarse feature maps from later layers embody rich semantic information. One can clearly identify the  semantic gap and uneven distribution of boundary information within this feature hierarchy. Albeit these limitations of CNN have been addressed to some extent via skip connections \cite{long2015fully,ronneberger2015u,badrinarayanan2015segnet,chen2018encoder}, repeatedly merging and transforming the feature maps across the feature hierarchy inevitably dilutes both the semantic and boundary information especially for fine-grained regions or objects such as body parts. 

To address the above issues, we propose a novel attentive decoder network, as illustrated in Fig. \ref{fig:net}, which consists of four layers (Layer 1, 2, 3, 4) corresponding to four encoder outputs. Each layer comprises of two branches, i.e., semantic and contour branches, to transform the feature maps to categorical (N-channel) space and contour (1-channel) space enabled by the deep supervision~\cite{lee2015deeply,wang2015training} of semantic labels (segmentation ground-truth) and contour maps (Canny edge map from segmentation ground-truth). Two \emph{element-wise} attention modules are proposed to process categorical and contour feature maps respectively. Each attention module is also connected across (illustrated as red and blue dashed arrows in Fig. \ref{fig:net}) the feature hierarchy to propagate consistent semantic and contour information in opposite directions respectively. 

Semantic attention module, as illustrated in Fig. \ref{fig:sem-boundary} (left), takes N-channel categorical feature maps from current layer (Layer i) and the N-channel processed feature map from later layer (Layer i+1) as inputs, and computes element-wise attention weights ($S_c$) by applying an element-wise sigmoid function on the summed feature maps, and weighs the feature map $F_c$ via $F'_c = S_c \odot F_c$. As a result the feature map of current layer is semantically enhanced by the feature map from later layer in categorical feature space. Similarly, contour attention module, as illustrated in Fig. \ref{fig:sem-boundary} (right), takes 1-channel contour feature maps from current layer (Layer i) and the 1-channel processed contour feature map from earlier layer (Layer i-1) as inputs, and computes element-wise attention weights ($S_b$) by applying an element-wise sigmoid function on the concatenated feature maps, and weighs the feature map $F_b$ via $F'_b = S_b \odot F_b$. As a result the contour feature of current layer is enhanced by earlier layer which has much stronger boundary information. Note, our attention mechanism is element-wise attention which is different from the channel-wise attention model proposed by \cite{yu2018learning}. Our element-wise attention model suits better for  semantic segmentation which is a dense prediction problem. Overall, there are consistent semantic and contour information flows across the CNN feature hierarchy which are missing in the state-of-the-art architectures. Feature maps from semantic and contour branches are summed in the \emph{attention fusion} module at each layer.

\subsection{Pose as Context}

In order to generate the initial person configuration, we employ Deeper-Cut \cite{insafutdinov2016deepercut}, which addresses the multi-person scenario explicitly using integer linear programming, to extract human skeleton map. We put forward a notion of using geodesic distance to encode the contextual information of people location and body part configuration. We employ superpixel maps  \cite{achanta2012slic}  rather than pixels to exploit pose context leveraging perceptually meaningful and coherent lower-level features. Specifically, we generate multi-layer superpixels with different granularities to encode the pose context at multiple scales. In each superpixel map, we compute the geodesic distance from all superpixels \wrt the set of superpixels associated with each skeleton line (corresponding to pose parameters $\Theta_p$). The geodesic distance $d(y_i , \Theta_p)$ between superpixel $y_i$ and skeleton line $\Theta_p$ is defined as the smallest integral of a weight function over all paths from $y_i$ to the set of superpixels $ \Omega_{p}$ associated with $\Theta_p$. Formally, $d_{geo}(y_i , \Theta_p)$ is computed as 
\begin{equation}
d_{geo}(y_i, \Theta_p) = \mymin _{y_j \in \Omega_{p}} d(y_i, y_j)  \nonumber
\end{equation}
where 
\begin{equation}
d(y_i, y_j)  = \mymin _{C_{y_i, y_j}} \int_{0}^{1} |W \cdot \dot{C}_{y_i,y_j} (s)| ds,   \nonumber
\end{equation}
where $C_{y_i, y_j}(s)$ is a path connecting the superpixels $y_i, y_j$ (for $s=0$ and $s=1$ respectively). The weights $W$ are set the gradient of the average CIE Lab color of superpixels, \ie $W = \nabla \bar{c}$ with $\bar{c}$ as the average color of superpixels. If a superpixel is outside the desired part, its geodesic distance is large since all the possible pathway to the skeleton line cross the boundaries with higher gradient value (edges) and spatial distance. Based on the geodesic distance, the likelihood that semantic part $\Theta_p$ occurs at superpixel $y_i$ can be computed as
\begin{equation}
p(y_i | \Theta_p) = \text{exp} (- \beta d_{geo}^2(y_i , \Theta_p))\nonumber
\end{equation}
where $\beta$ is a parameter which controls the decay of likelihood \wrt geodesic distance. 



\subsection{Graphical Model}

The computed superpixel based part confidence maps serve as the higher-level contexts of body configurations. Our goal is to combine this contextual information with the bottom-up predictions in a principled manner. To this end, we propose a novel graphical model to provide a unified framework for this task. 

We construct an undirected graph $G = (V, E)$ with pixels and superpixels as nodes $V = \{X, Y\}$ respectively. Denoted as $E = \{E_{\mathrm{XY}}, E_{\mathrm{XX}}, E_{\mathrm{YY}}\}$, we define the edges as links between pairs of graph nodes. The existence of edges is determined based on the spatial adjacency in the graph as follows.

\subsubsection{Pose-Pixel Edge $E_{\mathrm{XY}}$}

The connections between each superpixel and its constituent pixels are added as pose-pixel edges. Superpixels comprise of higher-level cues of human body configuration, as well as perceptually meaningful and coherent features.  The weight $w^{\mathrm{XY}}_{im} $ on edge $e^{\mathrm{XY}}_{im} \in E^{\mathrm{XY}}$ between pixel $x_i$ and superpixel $y_m$ is defined as $w^{\mathrm{XY}}_{im} =  [x_i \in y_m] \cdot e^{-(1-\mathbf{Pr}(x_i | y_m))}$,
where $[\cdot]$ is the indicator function, and $\mathbf{Pr}(x_i | y_m)$ is the likelihood of observing $x_i$ given the probability density function (PDF) of superpixel $y_m$. We estimate the probability density of superpixel via the fast kernel density estimation \cite{yang2003} on CIE Lab color. These pose-pixel edges transfer the person level multi-scale contextual cues into the pixels, whilst each superpixel can incorporate the complementary semantic information from local pixel. 

\subsubsection{Pixel Edge $E_{\mathrm{XX}}$}
All spatially adjacent (8-way connectivity neighborhood assumed) pixels are connected to form pixel edges $E_{\mathrm{XX}}$. 
Weight of an edge $e_{ij}^{\mathrm{XX}} \in E_{\mathrm{XX}}$ is defined to reflect both the local appearance similarity and spatial distance as follows. Let $\mathcal{N}_i$ be the set of pixels in the spatial neighborhood of $x_i$, we define $w^{\mathrm{XX}}_{ij} =   [ x_j \in \mathcal{N}_i] \cdot e^{-d^c(x_i, x_j)}$, 
where  $d^c(x_i, x_j)$ indicates the color distance between $x_i$ and $x_j$ which is defined as $d^{c}(i,j) = \frac{||c_i-c_j||^2}{2<||c_i-c_j||^2>}  $,
where $<\cdot>$ indicates expectation.

\subsubsection{Superpixel Edge $E_{\mathrm{YY}}$}

All spatially adjacent superpixels are connected to form superpixel edges $E_{\mathrm{YY}}$. Weight of an edge $e_{mn}^{\mathrm{YY}} \in E_{\mathrm{YY}}$ is defined to reflect both the local appearance similarity and spatial distance as follows. Let $\mathcal{N}_m^s$ be the set of superpixels in the spatial neighborhood of $y_m$.
\begin{equation}
w^{\mathrm{YY}}_{mn} =   [ y_n \in \mathcal{N}_m^s] \cdot e^{-\chi^2(h_m, h_n) d^s(y_m, y_n)}   \nonumber
\end{equation}
where  $\chi^2(h_m, h_n)$ is the $\chi^2$ distance between $L_1$-normalized CIE Lab color histograms $h_m$, $h_n$ of superpixels $y_m$ and $y_n$ respectively, and $d^s(y_m, y_n)$ indicates the spatial distance between $y_m$ and $y_n$.

\subsection{Cost Functions}

We define two novel quadratic cost functions to facilitate the joint inference of likelihoods for pixels and superpixels respectively,  harnessing the complementary contextual information to each other.

We infer the pixel likelihoods $\mathbf{U}_l$ by incorporating the superpixel likelihoods $\mathbf{V}_l$ as higher-level context cues in a principled manner. By characterizing the relationship between all nodes in the graph, the quadratic cost function $J_l^{\mathrm{X}}$ of pixel likelihoods $\mathbf{U}_l$ with respect to a label $l$ is as follows. Let the diagonal element of node degree matrix
$D^{\mathrm{X}} = \mathrm{diag}([d_1^{\mathrm{X}}, \dots, d_{N_{\mathrm{X}}}^{\mathrm{X}}])$ be defined as $d_i^{\mathrm{X}}=\sum_{j=1}^{N_{\mathrm{X}}} w_{ij}^{\mathrm{XX}}$.
\begin{align}
\label{eq:func1}
J_l^{\mathrm{X}} =&  J_{l, U}^{\mathrm{X}} + J_{l, P}^{\mathrm{X}} + J_{l, C}^{\mathrm{X}}  \nonumber \\
=&  \sum_{i=1}^{N_{\mathrm{X}}} \lambda^{\mathrm{X}} d_i^{\mathrm{X}} (u_{il} - \tilde{u}_{il})^2 + \sum_{i,j=1}^{N_{\mathrm{X}}} w_{ij}^{\mathrm{XX}} (u_{il} - u_{jl})^2 \nonumber \\
& + \sum_{i=1}^{N_{\mathrm{X}}} \pi d_i^{\mathrm{X}} (u_{il} - \bar{u}_{il})^2
\end{align}
where $\lambda$ and $\pi$ are parameters. The pixel probability $\tilde{u}_{il}$ is the initial likelihood with respect to label $l$ from the trained network. The estimated likelihood $\bar{u}_{il}$ of pixel $x_i$
from superpixel likelihood $v_{ml} \in \mathbf{V}_l$ is define as the weighted average of its corresponding superpixels from multi-layer superpixel maps,
\begin{equation}
\label{eq:uil}
\bar{u}_{il} = \sum_{m=1}^{N_{\mathrm{Y}}} p_{im}^{\mathrm{XY}} v_{ml}, 
\end{equation}
where 
\begin{equation}
p_{im}^{\mathrm{XY}} = \frac{w_{im}^{\mathrm{XY}}}{\sum_{m=1}^{N_{\mathrm{Y}}} w_{im}^{\mathrm{XY}}}.
\end{equation}
In this cost function, $J_{l, U}^{\mathrm{X}}$ and $J_{l, P}^{\mathrm{X}}$ are the fitting constraint and smoothness constraint respectively, while $J_{l, C}^{\mathrm{X}}$ is the contextual constraint.

$J_{l, U}^{\mathrm{X}}$ encourages
pixels to have the initial likelihood, which is controlled by $\lambda^{\mathrm{X}}$ measuring how much the inferred likelihood should agree with the initial likelihood. $J_{l, P}^{\mathrm{X}}$ promotes the continuity of inferred likelihood among adjacent nodes lying in a close vicinity in the feature space. $J_{l, C}^{\mathrm{X}}$ facilitates the inference of each pixel to be aware of higher-level context information. The intuition is that superpixel confidence map encodes richer semantics and intrinsic information of person parts and configuration, which can be propagated to its constituent pixels during inference.

In order to solve (\ref{eq:func1}), the superpixel likelihoods $\mathbf{V}_l$ are also required to be estimated by referring to the pixel likelihoods $\mathbf{U}_l$ in graph $G$. Similar to (\ref{eq:func1}), the cost function $J_l^{\mathrm{Y}}$ of superpixels likelihoods $\mathbf{V}_l$ is defined as follows. Let the diagonal element of node degree matrix
$D^{\mathrm{Y}} = \mathrm{diag}([d_1^{\mathrm{Y}}, \dots, d_{N_{\mathrm{Y}}}^{\mathrm{Y}}])$ be defined as  $d_m^{\mathrm{Y}}=\sum_{n=1}^{N_{\mathrm{Y}}} w_{mn}^{\mathrm{YY}}$,
\begin{align}
\label{eq:func2}
J_l^{\mathrm{Y}} =&  J_{l, U}^{\mathrm{Y}} + J_{l, P}^{\mathrm{Y}} + J_{l, C}^{\mathrm{Y}}  \nonumber \\
=&  \sum_{m=1}^{N_{\mathrm{Y}}} \lambda^{\mathrm{Y}} d_m^{\mathrm{Y}} (v_{ml} - \tilde{v}_{ml})^2 + \sum_{m,n=1}^{N_{\mathrm{Y}}} w_{mn}^{\mathrm{YY}} (v_{ml} - v_{nl})^2 \nonumber \\
&+ \sum_{m=1}^{N_{\mathrm{Y}}} \psi d_m^{\mathrm{Y}} (v_{ml} - \bar{v}_{ml})^2
\end{align}
where $\lambda^{\mathrm{Y}}$ and $\psi$ are parameters,  $\tilde{v}_{ml}$ is the initial likelihood of superpixels $m$ given label $l$ from the geodesic distance of skeleton notion, and the estimated likelihood $\bar{v}_{ml}$ of the superpixels $y_m$ is defined by incorporating local semantic cues, i.e., pixel likelihoods $\mathbf{U}_l$. $\bar{v}_{ml}$ is computed as the weighted average of its constituent pixel likelihoods:
\begin{equation}
\bar{v}_{ml} = \sum_{i=1}^{N_{\mathrm{X}}} p_{mi}^{\mathrm{YX}} u_{il}, \nonumber
\end{equation}
where 
\begin{equation}
p_{mi}^{\mathrm{YX}} = \frac{w_{mi}^{\mathrm{YX}}}{\sum_{i=1}^{N_{\mathrm{X}}} w_{mi}^{\mathrm{YX}}}.
\end{equation}

Similarly, (\ref{eq:func2}) consists of three terms, i.e., $J_{l, U}^{\mathrm{Y}}$, $J_{l, P}^{\mathrm{Y}}$ and $J_{l, C}^{\mathrm{Y}}$. $J_{l, U}^{\mathrm{Y}}$ is the fitting constraint that encourages each superpixels to have its initial likelihood. $J_{l, P}^{\mathrm{Y}}$ is the smoothness constraint which promotes label continuity among superpixels to preserve the configuration of person pose. The third term $J_{l, C}^{\mathrm{Y}}$ is the contextual constraint which collects local semantic cues in a bottom-up manner to refine the superpixel likelihood $\mathbf{V}_l$ using pixel likelihoods $\mathbf{U}_l$, since it can not guarantee that the estimated pose is always correct and the superpixel might carry erroneous contextual information.

\subsection{Convex Optimization}

Two cost functions $\mathbf{U}_l$ and $\mathbf{V}_l$ should be minimized simultaneously since they are complementary to each other. We reformulate these functions as matrix forms with respect to the likelihoods $\mathbf{U}_l = [u_{il}]_{N_{\mathrm{X}} \times 1}$ and $\mathbf{V}_l = [v_{ml}]_{N_{\mathrm{Y}} \times 1}$ from the initial likelihoods $\tilde{\mathbf{U}}_l = [\tilde{u}_{il}]_{N_{\mathrm{X}} \times 1}$ and $\tilde{\mathbf{V}}_l = [\tilde{v}_{ml}]_{N_{\mathrm{Y}} \times 1}$  respectively,
\begin{align}
J_l^{\mathrm{X}} = & (\mathbf{U}_l - \tilde{\mathbf{U}}_l) ^{T} \mathbf{D}^{\mathrm{X}} \mathbf{\Lambda}^{\mathrm{X}} (\mathbf{U}_l - \tilde{\mathbf{U}}_l)  + \mathbf{U}_l^T(\mathbf{D}^{\mathrm{X}} - \mathbf{W}^{\mathrm{X}}) \mathbf{U}_l    \nonumber \\
& + \pi (\mathbf{U}_l - \mathbf{P}^{\mathrm{XY}} \mathbf{V}_l)^{T}  D^{\mathrm{X}} (\mathbf{U}_l - \mathbf{P}^{\mathrm{XY}} \mathbf{V}_l) \\ \nonumber
J_l^{\mathrm{Y}} = & (\mathbf{V}_l - \tilde{\mathbf{V}}_l) ^{T} \mathbf{D}^{\mathrm{Y}} \mathbf{\Lambda}^{\mathrm{Y}} (\mathbf{V}_l - \tilde{\mathbf{V}}_l)  + \mathbf{V}_l^T(\mathbf{D}^{\mathrm{Y}} - \mathbf{W}^{\mathrm{Y}}) \mathbf{V}_l    \nonumber \\
& + \psi (\mathbf{V}_l - \mathbf{P}^{\mathrm{YX}} \mathbf{U}_l)^{T}  D^{\mathrm{Y}} (\mathbf{V}_l - \mathbf{P}^{\mathrm{YX}} \mathbf{U}_l) \\ \nonumber
\end{align}
where $\mathbf{W}^{\mathrm{X}} = [w^{\mathrm{XX}}_{ij}]_{N_{\mathrm{X}} \times N_{\mathrm{X}}}$ and $\mathbf{W}^{\mathrm{Y}} = [w^{\mathrm{YY}}_{mn}]_{N_{\mathrm{Y}} \times N_{\mathrm{Y}}}$. The contextual dependencies between pixels and their corresponding superpixels in graph $G$ are formulated by $\mathbf{P}^{\mathrm{XY}} = [p^{\mathrm{XY}}_{im}]_{N_{\mathrm{X}} \times N_{\mathrm{Y}}}$ and $\mathbf{P}^{\mathrm{YX}} = [p^{\mathrm{YX}}_{mi}]_{N_{\mathrm{Y}} \times N_{\mathrm{X}}}$. The diagonal elements of $N_{\mathrm{X}} \times N_{\mathrm{X}}$ matrix $\mathbf{\Lambda}^{\mathrm{X}} = \mathrm{diag}([\lambda^{\mathrm{X}}, \cdots, \lambda^{\mathrm{X}}])$ and $N_{\mathrm{Y}} \times N_{\mathrm{Y}}$ matrix $\mathbf{\Lambda}^{\mathrm{Y}} = \mathrm{diag}([\lambda^{\mathrm{Y}}, \cdots, \lambda^{\mathrm{Y}}])$ are the parameter $\lambda^{\mathrm{X}}$ and $\lambda^{\mathrm{X}}$ respectively.

By differentiating $J_l^{\mathrm{X}}$ and $J_l^{\mathrm{Y}}$ with respect to $\mathbf{U}_l$ and $\mathbf{V}_l$ respectively, we have
\begin{align}
\label{eq:pjx}
\frac{\partial J_l^{\mathrm{X}}}{\partial \mathbf{U}_l} &= \mathbf{U}_l  (\mathbf{I}^{\mathrm{X}}- \mathbf{P}^{\mathrm{X}}) + \mathbf{\Lambda}^{\mathrm{X}} (\mathbf{U}_l - \tilde{\mathbf{U}}_l) \nonumber \\
& + \pi (\mathbf{U}_l - \mathbf{P}^{\mathrm{XY}} \mathbf{V}_l) = 0   \\
\label{eq:pjy}
\frac{\partial J_l^{\mathrm{Y}}}{\partial \mathbf{V}_l} &= \mathbf{V}_l  (\mathbf{I}^{\mathrm{Y}}- \mathbf{P}^{\mathrm{Y}}) + \mathbf{\Lambda}^{\mathrm{Y}} (\mathbf{V}_l - \tilde{\mathbf{V}}_l) \nonumber\\
& + \psi (\mathbf{V}_l - \mathbf{P}^{\mathrm{YX}} \mathbf{U}_l) = 0
\end{align}
where $\mathbf{P}^{\mathrm{X}} = {\mathbf{D}^{\mathrm{X}}}^{-1} \mathbf{W}^{\mathrm{X}}$ (or $\mathbf{P}^{\mathrm{Y}} = {\mathbf{D}^{\mathrm{Y}}}^{-1} \mathbf{W}^{\mathrm{Y}}$), and $\mathbf{I}^{\mathrm{X}}$ (or $\mathbf{I}^{\mathrm{Y}}$) is identify matrix.

By denoting all likelihoods as $\mathbf{Z}_l = [\mathbf{U}_l; \mathbf{V}_l]$ and initial likelihoods as $\tilde{\mathbf{Z}}_l = [\tilde{\mathbf{U}}_l; \tilde{\mathbf{V}}_l]$, (\ref{eq:pjx}) and (\ref{eq:pjy}) can be jointly transformed into
\begin{equation}
\label{eq:lineq}
(\mathbf{I} - (\mathbf{I} - \Gamma) \Pi) \mathbf{Z}_l  =  \Gamma \tilde{\mathbf{Z}},
\end{equation}
where
\begin{equation}
\Pi = \begin{bmatrix} \frac{\mathbf{P}^{\mathrm{X}}}{1+\pi} & \frac{\pi\mathbf{P}^{\mathrm{XY}}}{1+\pi} \\ \frac{\psi\mathbf{P}^{\mathrm{YX}}}{1+\psi} & \frac{\mathbf{P}^{\mathrm{Y}}}{1+\psi} \end{bmatrix},
\end{equation}
and
\begin{equation}
\Gamma = \begin{bmatrix} \frac{\mathbf{\Lambda}^{\mathrm{X}}}{(1+\pi)\mathbf{I} + \mathbf{\Lambda}^{\mathrm{X}}} &  \\  & \frac{\mathbf{\Lambda}^{\mathrm{Y}}}{(1+\psi)\mathbf{I} + \mathbf{\Lambda}^{\mathrm{Y}}} \end{bmatrix}.
\end{equation}
Denoting 
\begin{equation}
\mathbf{B} = \mathbf{I} - (\mathbf{I} - \Gamma) \Pi, 
\end{equation}
(\ref{eq:lineq}) can be solved by a sparse matrix inversion
\begin{equation}
\mathbf{Z}_l  =  \mathbf{B}^{-1} \Gamma \tilde{\mathbf{Z}}.
\end{equation}


\subsection{Semantic Part Labeling}
Posterior probabilities of each pixel with respect to part label $l$ can then be computed following Bayes rule
\begin{equation}
p(l | x_i) =  \frac{p(x_i | l) p(l)}{\sum_{l^{\prime} = 1}^L p(x_i | l^{\prime}) p(l^{\prime})} =  \frac{u_{il}}{\sum_{l^{\prime} = 1}^L u_{il^{\prime}}}
\end{equation}
where an uniform prior probability $p(l)$ is assumed.

Each pixel is finally assigned with the label corresponding to the class with the \emph{maximum a posterior} probability, which constitutes to the semantic person part segmentation,
\begin{equation}
\hat{l}_i =\myargmax_l p(l | x_i)
\end{equation}

\begin{table*}[t!]
	\caption{Quantitatively segmentation results on Pascal Person-Part dataset}
	\centering 
	\small
	\begin{tabular}{c|cccccccc} 
		\toprule
		Method & Head & Torso & U-arms  & L-arms & U-legs & L-legs & Background & Avg.\\
		\midrule
		DeepLab-LargeFOV \cite{ChenPKMY14}  & 78.09 &  54.02 &   37.29 & 36.85  & 33.73 & 29.61 & 92.85 & 51.78\\
		HAZN \cite{xia2016zoom}     & 80.79 & 59.11 & 43.05 &  42.76 & 38.99 & 34.46 & 93.59 & 56.11  \\
		Attention \cite{chen2016attention}  & - & - & - & - &  -  &  - & - & 56.39\\
		LG-LSTM \cite{liang2016semantic2}    & 82.72 & 60.99 & 45.40 & 47.76  & 42.33 & 37.96 & 88.63 & 57.97 \\
		Graph LSTM \cite{liang2016semantic} & 82.69 & 62.68 & 46.88 & 47.71  & 45.66 & 40.93 & 94.59 & 60.16 \\
		DeepLab v2 \cite{chen2016deeplab} & - & - & - & - &  -  &  - & - & 58.90\\
		JPS  (final, CRF) \cite{xia2017joint} & 85.50 & 67.87 & 54.72 & 54.30 & 48.25 & 44.76 & 95.32 & 64.39 \\
		PCNet-126 \cite{ZhuCTW18} & 86.81 & 69.06 & 55.35 & 55.27 & 50.21 & 48.54 &96.07 & 65.90\\
		\midrule
		Our model (w/o graph) & 89.19 & 74.88 & 55.98 & 60.76  & 50.76 & 41.45 & 95.12 & 66.87 \\
		Our model (final)  &90.84 &75.85 & 56.18 & 64.86 & 52.86 & 43.52 & 95.75 & \textbf{68.55} \\
		\bottomrule
	\end{tabular} \label{tbl:pascal-result} 
\end{table*}

\begin{figure*}[ht!]
	\centering
	\includegraphics[width=0.99\linewidth]{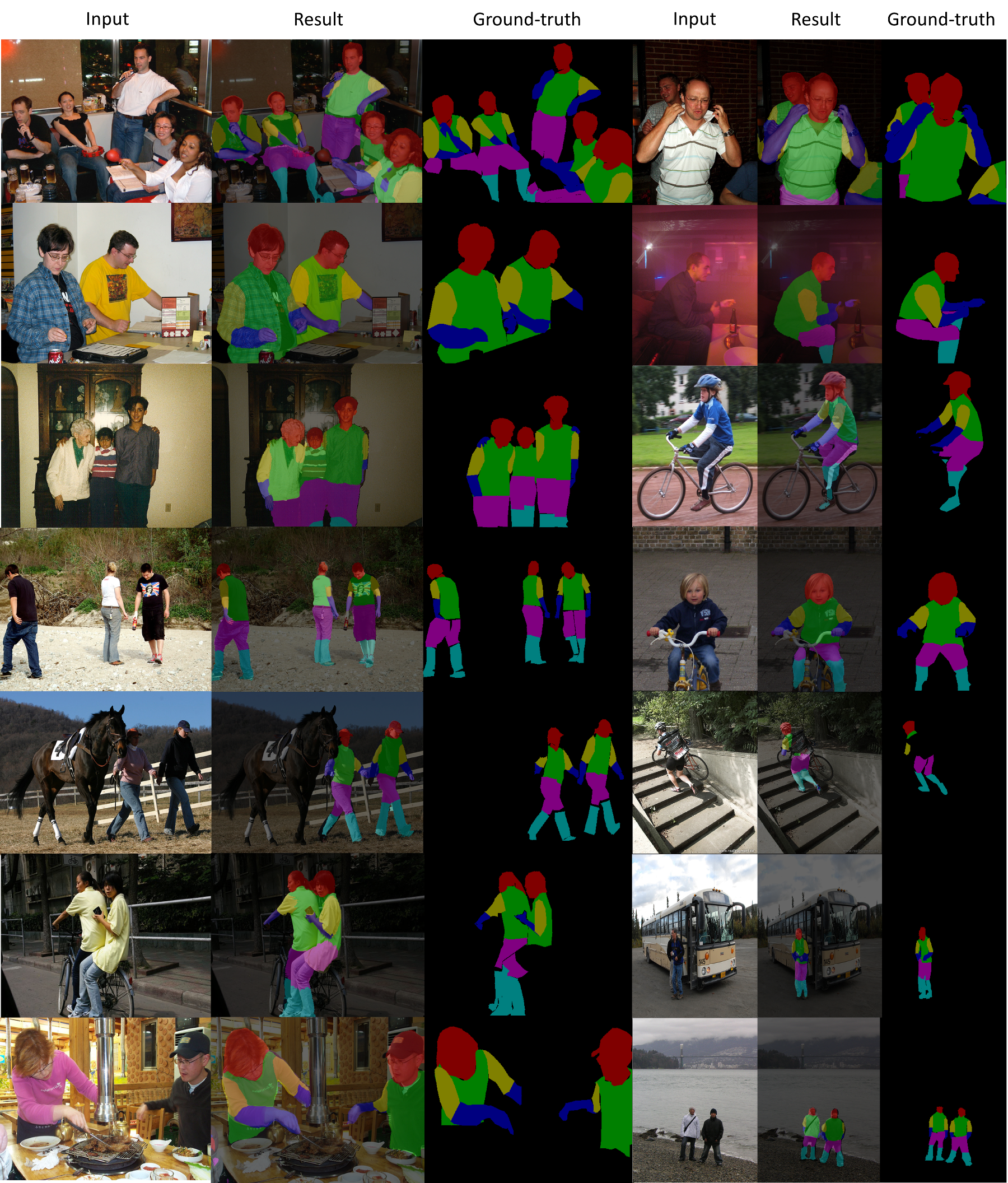}
	\caption{Qualitative results of our algorithm on Pascal Person-Part dataset.  \label{fig:pascal}}
\end{figure*}

\section{Experiments}
\label{sec:evaluation}

We evaluate our proposed approach on the Pascal Person-Part dataset \cite{chen2014detect} which is a subset from the PASCAL VOC 2010 dataset, containing 1716 training images, and 1817 test images. This dataset contains multiple person per image in unconstrained poses and challenging natural environments with detailed person part annotations (head, torso, upper/lower arms and upper/lower legs) and one background class. The segmentation results are evaluated in terms of mean Intersection-over-Union (mIOU) averaged across classes. We adopt ResNet-101 pretrained on ImageNet as encoder.

As shown in Tab. \ref{tbl:pascal-result}, we compare our model with 7 state-of-the-art methods, \ie DeepLab-LargeFOV \cite{ChenPKMY14}, HAZN \cite{xia2016zoom}, Attention \cite{chen2016attention}, LG-LSTM \cite{liang2016semantic2}, Graph LSTM \cite{liang2016semantic}, DeepLab v2 \cite{chen2016deeplab} and JPS \cite{xia2017joint}. Our proposed approach
obtains the performance of $68.55\%$, which outperforms all the compared methods. Our approach improves the segmentation accuracy in all parts with considerable margins. Comparing with DeepLab v2 which also uses ResNet-101 as encoder, our architecture significantly improves the accuracy by $7.97\%$, which demonstrates the capability of our network in dealing with fine-grained body parts. As ablation study, we evaluate our architecture in comparison to a baseline network without semantic and contour attention modules. The result shows that incorporating semantic attention module improves the performance by $2.43\%$. We further observe a performance improvement of $1.36\%$ by introducing the contour attention module. Our architecture outperfoms the state-of-the-art end-to-end body parsing network PCNet-126 \cite{ZhuCTW18} by ~1\%.

All top-down approaches, \ie JPS and ours, significantly surpass bottom-up approaches DeepLab-LargeFOV (w/o CRF), HAZN, Attention, LG-LSTM, Graph LSTM and DeepLab-v2 (w/o CRF) with large margins. Bottom-up approaches normally rely on pixel level training and inference without explicitly taking into account of person part configuration, and thus they are error-prone to unusual human poses, as well as scenes where people are overlapping and interacting.  This comparison confirms the benefits of incorporating person-level information in resolving the ambiguity of various parts in challenging situations.  

Our method outperforms the best competing method JPS \cite{xia2017joint} by $4.16\%$. Note we do not use any CRF for post processing. 
JPS takes a similar top-down approach as our method, however it directly utilizes skeletons from estimated pose as confidence map to improve the part segmentation which is sensitive to erroneous person pose estimations. 
Our approach is able to systematically integrate the higher-level part configuration information and local semantic cues into a graphical model, which recursively estimates both to propagate long-range intrinsic structure to facilitate local semantic labeling against spurious pose context and pixel level predictions. 

More qualitative evaluations on Pascal Person-Part dataset are provided in Fig. \ref{fig:pascal}. Specifically, for multi-person highly overlapping instances (\eg rows 1 and 6), our method can resolve the inter-person occlusions, relieve the local inter-part confusions (\eg lifted lower arms in row 1), recover the parts for small scale human instances (\eg right images of rows 6 and 7) and exclude regions occluded from non-person object (\eg row 5).

\section{Conclusion}

We have proposed a network architecture which comprises of novel semantic and contour attention mechanisms to resolve the semantic ambiguities and boundary localization issues related to fine-grained body parsing problem. We further proposed a novel graphical model to tackle the multi-person semantic part segmentation problem in challenging natural scenes. This model is able to combine person level configuration information with the pixel level semantic cues in a unified framework, where the lower-level semantic cues can be recursively estimated by propagating higher-level contextual information from estimated pose and vice versa, across the proposed graph. We further proposed an optimization technique to efficiently estimate the cost functions. The experiments confirmed the significantly improved robustness and accuracy over the state-of-the-art approaches on challenging Pascal Person-Part dataset.

\bibliographystyle{splncs03}
\bibliography{refs}

\begin{thebibliography}{10}
\providecommand{\url}[1]{\texttt{#1}}
\providecommand{\urlprefix}{URL }

\bibitem{achanta2012slic}
Achanta, R., Shaji, A., Smith, K., Lucchi, A., Fua, P., S{\"u}sstrunk, S.: Slic
  superpixels compared to state-of-the-art superpixel methods. IEEE Trans.
  Pattern Anal. Mach. Intell.  34(11),  2274--2282 (2012)

\bibitem{badrinarayanan2015segnet}
Badrinarayanan, V., Kendall, A., Cipolla, R.: Segnet: A deep convolutional
  encoder-decoder architecture for image segmentation. arXiv preprint
  arXiv:1511.00561  (2015)

\bibitem{ChenPKMY14}
Chen, L., Papandreou, G., Kokkinos, I., Murphy, K., Yuille, A.L.: Semantic
  image segmentation with deep convolutional nets and fully connected crfs.
  CoRR  abs/1412.7062 (2014)

\bibitem{chen2016deeplab}
Chen, L.C., Papandreou, G., Kokkinos, I., Murphy, K., Yuille, A.L.: Deeplab:
  Semantic image segmentation with deep convolutional nets, atrous convolution,
  and fully connected crfs. arXiv preprint arXiv:1606.00915  (2016)

\bibitem{chen2016attention}
Chen, L.C., Yang, Y., Wang, J., Xu, W., Yuille, A.L.: Attention to scale:
  Scale-aware semantic image segmentation. In: CVPR. pp. 3640--3649 (2016)

\bibitem{chen2018encoder}
Chen, L.C., Zhu, Y., Papandreou, G., Schroff, F., Adam, H.: Encoder-decoder
  with atrous separable convolution for semantic image segmentation. arXiv
  preprint arXiv:1802.02611  (2018)

\bibitem{chen2020fine}
Chen, S., Zhao, Y., Jin, Q., Wu, Q.: Fine-grained video-text retrieval with
  hierarchical graph reasoning. In: Proceedings of the IEEE/CVF conference on
  computer vision and pattern recognition. pp. 10638--10647 (2020)

\bibitem{chen2014detect}
Chen, X., Mottaghi, R., Liu, X., Fidler, S., Urtasun, R., Yuille, A.: Detect
  what you can: Detecting and representing objects using holistic models and
  body parts. In: CVPR. pp. 1971--1978 (2014)

\bibitem{deng2021generative}
Deng, F., Zhi, Z., Lee, D., Ahn, S.: Generative scene graph networks. In:
  International Conference on Learning Representations (2021)

\bibitem{hu2013markov}
Hu, R., James, S., Wang, T., Collomosse, J.: Markov random fields for sketch
  based video retrieval. In: Proceedings of the 3rd ACM conference on
  International conference on multimedia retrieval. pp. 279--286 (2013)

\bibitem{wang11}
Hu, R., Wang, T., Collomosse, J.P.: A bag-of-regions approach to sketch-based
  image retrieval. In: ICIP. pp. 3661--3664 (2011)

\bibitem{insafutdinov2016deepercut}
Insafutdinov, E., Pishchulin, L., Andres, B., Andriluka, M., Schiele, B.:
  Deepercut: A deeper, stronger, and faster multi-person pose estimation model.
  In: ECCV. pp. 34--50 (2016)

\bibitem{jiang2016detangling}
Jiang, H., Grauman, K.: Detangling people: Individuating multiple close people
  and their body parts via region assembly. arXiv preprint arXiv:1604.03880
  (2016)

\bibitem{kalayeh2018human}
Kalayeh, M.M., Basaran, E., G{\"o}kmen, M., Kamasak, M.E., Shah, M.: Human
  semantic parsing for person re-identification. In: Proceedings of the IEEE
  Conference on Computer Vision and Pattern Recognition. pp. 1062--1071 (2018)

\bibitem{kyprianidis2012state}
Kyprianidis, J.E., Collomosse, J., Wang, T., Isenberg, T.: State of the"
  art”: A taxonomy of artistic stylization techniques for images and video.
  IEEE transactions on visualization and computer graphics  19(5),  866--885
  (2012)

\bibitem{lee2015deeply}
Lee, C.Y., Xie, S., Gallagher, P., Zhang, Z., Tu, Z.: Deeply-supervised nets.
  In: Artificial Intelligence and Statistics. pp. 562--570 (2015)

\bibitem{li2018pose}
Li, D., Chen, X., Zhang, Z., Huang, K.: Pose guided deep model for pedestrian
  attribute recognition in surveillance scenarios. In: 2018 IEEE International
  Conference on Multimedia and Expo (ICME). pp. 1--6. IEEE (2018)

\bibitem{li2017holistic}
Li, Q., Arnab, A., Torr, P.H.: Holistic, instance-level human parsing. arXiv
  preprint arXiv:1709.03612  (2017)

\bibitem{li2018diversity}
Li, S., Bak, S., Carr, P., Wang, X.: Diversity regularized spatiotemporal
  attention for video-based person re-identification. In: Proceedings of the
  IEEE Conference on Computer Vision and Pattern Recognition. pp. 369--378
  (2018)

\bibitem{liang2016semantic}
Liang, X., Shen, X., Feng, J., Lin, L., Yan, S.: Semantic object parsing with
  graph lstm. In: ECCV. pp. 125--143. Springer (2016)

\bibitem{liang2016semantic2}
Liang, X., Shen, X., Xiang, D., Feng, J., Lin, L., Yan, S.: Semantic object
  parsing with local-global long short-term memory. In: CVPR. pp. 3185--3193
  (2016)

\bibitem{long2015fully}
Long, J., Shelhamer, E., Darrell, T.: Fully convolutional networks for semantic
  segmentation. In: CVPR. pp. 3431--3440 (2015)

\bibitem{lu2020video}
Lu, X., Wang, W., Danelljan, M., Zhou, T., Shen, J., Van~Gool, L.: Video object
  segmentation with episodic graph memory networks. In: Computer Vision--ECCV
  2020: 16th European Conference, Glasgow, UK, August 23--28, 2020,
  Proceedings, Part III 16. pp. 661--679. Springer (2020)

\bibitem{luo2013pedestrian}
Luo, P., Wang, X., Tang, X.: Pedestrian parsing via deep decompositional
  network. In: ICCV. pp. 2648--2655 (2013)

\bibitem{ma2011human}
Ma, L., Yang, X., Xu, Y., Zhu, J.: Human identification using body prior and
  generalized emd. In: ICIP. pp. 1441--1444. IEEE (2011)

\bibitem{qi20173d}
Qi, X., Liao, R., Jia, J., Fidler, S., Urtasun, R.: 3d graph neural networks
  for rgbd semantic segmentation. In: Proceedings of the IEEE international
  conference on computer vision. pp. 5199--5208 (2017)

\bibitem{ronneberger2015u}
Ronneberger, O., Fischer, P., Brox, T.: U-net: Convolutional networks for
  biomedical image segmentation. In: International Conference on Medical image
  computing and computer-assisted intervention. pp. 234--241. Springer (2015)

\bibitem{WangRLWK16}
Wang, H., Raiko, T., Lensu, L., Wang, T., Karhunen, J.: Semi-supervised domain
  adaptation for weakly labeled semantic video object segmentation. In: Asian
  Conference on Computer Vision. pp. 163--179 (2016)

\bibitem{wang2016semi}
Wang, H., Raiko, T., Lensu, L., Wang, T., Karhunen, J.: Semi-supervised domain
  adaptation for weakly labeled semantic video object segmentation. In: ACCV.
  pp. 163--179 (2016)

\bibitem{WangW16}
Wang, H., Wang, T.: Boosting objectness: Semi-supervised learning for object
  detection and segmentation in multi-view images. In: ICASSP. pp. 1796--1800
  (2016)

\bibitem{wang2016primary}
Wang, H., Wang, T.: Primary object discovery and segmentation in videos via
  graph-based transductive inference. Computer Vision and Image Understanding
  143,  159--172 (2016)

\bibitem{wang2017}
Wang, H., Wang, T., Chen, K., K{\"a}m{\"a}r{\"a}inen, J.K.: Cross-granularity
  graph inference for semantic video object segmentation. In: IJCAI. pp.
  4544--4550 (2017)

\bibitem{wang2017cross}
Wang, H., Wang, T., Chen, K., K{\"a}m{\"a}r{\"a}inen, J.K.: Cross-granularity
  graph inference for semantic video object segmentation. In: IJCAI. pp.
  4544--4550 (2017)

\bibitem{wang2015training}
Wang, L., Lee, C.Y., Tu, Z., Lazebnik, S.: Training deeper convolutional
  networks with deep supervision. arXiv preprint arXiv:1505.02496  (2015)

\bibitem{wang2015joint}
Wang, P., Shen, X., Lin, Z., Cohen, S., Price, B., Yuille, A.L.: Joint object
  and part segmentation using deep learned potentials. In: ICCV. pp. 1573--1581
  (2015)

\bibitem{wang2021end}
Wang, T., Xu, N., Chen, K., Lin, W.: End-to-end video instance segmentation via
  spatial-temporal graph neural networks. In: Proceedings of the IEEE/CVF
  International Conference on Computer Vision. pp. 10797--10806 (2021)

\bibitem{tinghuai2016apparatus}
Wang, T.: Apparatus, a method and a computer program for image processing
  (Nov~15 2016), uS Patent 9,495,755

\bibitem{tinghuai2017method}
Wang, T.: Method, apparatus and computer program product for segmentation of
  objects in media content (Apr~25 2017), uS Patent 9,633,446

\bibitem{wang2017submodular}
Wang, T.: Submodular video object proposal selection for semantic object
  segmentation. In: 2017 IEEE International Conference on Image Processing
  (ICIP). pp. 4522--4526. IEEE (2017)

\bibitem{tinghuai2018method1}
Wang, T.: Method for analysing media content (Apr~26 2018), uS Patent App.
  15/785,711

\bibitem{tinghuai2020semantic}
Wang, T.: Semantic segmentation based on a hierarchy of neural networks (Dec~22
  2020), uS Patent 10,872,275

\bibitem{wang2014wide}
Wang, T., Collomosse, J., Hilton, A.: Wide baseline multi-view video matting
  using a hybrid markov random field. In: 2014 22nd International Conference on
  Pattern Recognition. pp. 136--141. IEEE (2014)

\bibitem{wang2011stylized}
Wang, T., Collomosse, J., Hu, R., Slatter, D., Greig, D., Cheatle, P.: Stylized
  ambient displays of digital media collections. Computers \& Graphics  35(1),
  54--66 (2011)

\bibitem{wang2010video}
Wang, T., Collomosse, J., Slatter, D., Cheatle, P., Greig, D.: Video
  stylization for digital ambient displays of home movies. In: Proceedings of
  the 8th International Symposium on Non-Photorealistic Animation and
  Rendering. pp. 137--146 (2010)

\bibitem{wang2013learnable}
Wang, T., Collomosse, J.P., Hunter, A., Greig, D.: Learnable stroke models for
  example-based portrait painting. In: BMVC (2013)

\bibitem{WangCSCG10}
Wang, T., Collomosse, J.P., Slatter, D., Cheatle, P., Greig, D.: Video
  stylization for digital ambient displays of home movies. In: NPAR. pp.
  137--146 (2010)

\bibitem{tinghuai2020watermark}
Wang, T., Fan, L.: Watermark embedding techniques for neural networks and their
  use (Jan~16 2020), uS Patent App. 16/508,434

\bibitem{wang2010multi}
Wang, T., Guillemaut, J.Y., Collomosse, J.: Multi-label propagation for
  coherent video segmentation and artistic stylization. In: 2010 IEEE
  International Conference on Image Processing. pp. 3005--3008. IEEE (2010)

\bibitem{WangHC14}
Wang, T., Han, B., Collomosse, J.P.: Touchcut: Fast image and video
  segmentation using single-touch interaction. Computer Vision and Image
  Understanding  120,  14--30 (2014)

\bibitem{wang2020spectral}
Wang, T., Wang, G., Tan, K.E., Tan, D.: Spectral pyramid graph attention
  network for hyperspectral image classification. arXiv preprint
  arXiv:2001.07108  (2020)

\bibitem{wang2014graph}
Wang, T., Wang, H.: Graph transduction learning of object proposals for video
  object segmentation. In: ACCV. pp. 553--568 (2014)

\bibitem{WangW14}
Wang, T., Wang, H.: Graph transduction learning of object proposals for video
  object segmentation. In: ACCV. pp. 553--568 (2014)

\bibitem{tinghuai2016method}
Wang, T., Wang, H.: Method and an apparatus for automatic segmentation of an
  object (May~5 2016), uS Patent App. 14/930,392

\bibitem{WangW18}
Wang, T., Wang, H.: Non-parametric contextual relationship learning for
  semantic video object segmentation. In: Progress in Pattern Recognition,
  Image Analysis, Computer Vision, and Applications - 23rd Iberoamerican
  Congress, {CIARP} 2018. pp. 325--333 (2018)

\bibitem{wang2019graph}
Wang, T., Wang, H.: Graph-boosted attentive network for semantic body parsing.
  In: Artificial Neural Networks and Machine Learning--ICANN 2019: Image
  Processing: 28th International Conference on Artificial Neural Networks,
  Munich, Germany, September 17--19, 2019, Proceedings, Part III 28. pp.
  267--280. Springer (2019)

\bibitem{wang2015robust}
Wang, T., Wang, H., Fan, L.: Robust interactive image segmentation with weak
  supervision for mobile touch screen devices. In: Proceedings of International
  Conference on Multimedia and Expo. pp. 1--6. IEEE (2015)

\bibitem{wang2015weakly}
Wang, T., Wang, H., Fan, L.: A weakly supervised geodesic level set framework
  for interactive image segmentation. Neurocomputing  168,  55--64 (2015)

\bibitem{wang2019zero}
Wang, W., Lu, X., Shen, J., Crandall, D.J., Shao, L.: Zero-shot video object
  segmentation via attentive graph neural networks. In: Proceedings of the
  IEEE/CVF international conference on computer vision. pp. 9236--9245 (2019)

\bibitem{wei2017glad}
Wei, L., Zhang, S., Yao, H., Gao, W., Tian, Q.: Glad: Global-local-alignment
  descriptor for pedestrian retrieval. In: Proceedings of the 25th ACM
  international conference on Multimedia. pp. 420--428. ACM (2017)

\bibitem{xia2016zoom}
Xia, F., Wang, P., Chen, L.C., Yuille, A.L.: Zoom better to see clearer: Human
  and object parsing with hierarchical auto-zoom net. In: ECCV. pp. 648--663.
  Springer (2016)

\bibitem{xia2017joint}
Xia, F., Wang, P., Chen, X., Yuille, A.: Joint multi-person pose estimation and
  semantic part segmentation. arXiv preprint arXiv:1708.03383  (2017)

\bibitem{xing2021learning}
Xing, Y., He, T., Xiao, T., Wang, Y., Xiong, Y., Xia, W., Wipf, D., Zhang, Z.,
  Soatto, S.: Learning hierarchical graph neural networks for image clustering.
  In: Proceedings of the IEEE/CVF International Conference on Computer Vision.
  pp. 3467--3477 (2021)

\bibitem{yamaguchi2015retrieving}
Yamaguchi, K., Kiapour, M.H., Ortiz, L.E., Berg, T.L.: Retrieving similar
  styles to parse clothing. IEEE Trans. Pattern Anal. Mach. Intell.  37(5),
  1028--1040 (2015)

\bibitem{yang2003}
Yang, C., Duraiswami, R., Gumerov, N.A., Davis, L.: Improved fast gauss
  transform and efficient kernel density estimation. In: ICCV. p. 464 (2003)

\bibitem{yang2021learning}
Yang, Y., Ren, Z., Li, H., Zhou, C., Wang, X., Hua, G.: Learning dynamics via
  graph neural networks for human pose estimation and tracking. In: Proceedings
  of the IEEE/CVF conference on computer vision and pattern recognition. pp.
  8074--8084 (2021)

\bibitem{yu2018learning}
Yu, C., Wang, J., Peng, C., Gao, C., Yu, G., Sang, N.: Learning a
  discriminative feature network for semantic segmentation. arXiv preprint
  arXiv:1804.09337  (2018)

\bibitem{zhang2014part}
Zhang, N., Donahue, J., Girshick, R., Darrell, T.: Part-based r-cnns for
  fine-grained category detection. In: ECCV. pp. 834--849. Springer (2014)

\bibitem{zhao2021graphfpn}
Zhao, G., Ge, W., Yu, Y.: Graphfpn: Graph feature pyramid network for object
  detection. In: Proceedings of the IEEE/CVF international conference on
  computer vision. pp. 2763--2772 (2021)

\bibitem{ZhuCTW18}
Zhu, B., Chen, Y., Tang, M., Wang, J.: Progressive cognitive human parsing. In:
  AAAI. pp. 7607--7614 (2018)

\bibitem{zhu2019cross}
Zhu, L., Wang, T., Aksu, E., Kamarainen, J.K.: Cross-granularity attention
  network for semantic segmentation. In: Proceedings of the IEEE/CVF
  International Conference on Computer Vision Workshops. pp. 0--0 (2019)

\bibitem{ZhuWAK19}
Zhu, L., Wang, T., Aksu, E., Kamarainen, J.: Portrait instance segmentation for
  mobile devices. In: ICME. pp. 1630--1635 (2019)

\end{thebibliography}
\end{document}